\documentclass[sigconf]{acmart}

\AtBeginDocument{%
  }

\copyrightyear{2026}
\acmYear{2026}
\setcopyright{cc}
\setcctype{by}
\acmConference[MM '26]{Proceedings of the 34th ACM International Conference on Multimedia}{November 10--14, 2026}{Rio de Janeiro, Brazil}
\acmBooktitle{Proceedings of the 34th ACM International Conference on Multimedia (MM '26), November 10--14, 2026, Rio de Janeiro, Brazil}
\acmDOI{10.1145/3767308.3839003}
\acmISBN{979-8-4007-2213-4/2026/11}

\usepackage{listings}
\usepackage{subcaption}
\usepackage{tikz}
\usetikzlibrary{arrows.meta,fit,positioning,calc,backgrounds}
\usepackage{booktabs}
\usepackage{multirow}
\usepackage{pifont}
\usepackage{xcolor}
\usepackage{float}

\newcommand{\method}{TSRN-RTVD}

\definecolor{brandgreen}{HTML}{2CA25F}

\begin{document}

\title{TSRN-RTVD: Real-Time Video Deblurring System}

\author{Nikita Alutis}
\orcid{0009-0007-6825-9720}
\affiliation{%
  \institution{Lomonosov Moscow State University}
  \city{Moscow}
  \country{Russian Federation}}
\email{nikita.alutis@graphics.cs.msu.ru}

\author{Danila Evsyukov}
\orcid{0009-0000-9714-2256}
\affiliation{%
  \institution{Lomonosov Moscow State University}
  \city{Moscow}
  \country{Russian Federation}}
\email{danila.evsyukov@graphics.cs.msu.ru}

\author{Egor Chistov}
\orcid{0009-0004-6156-195X}
\affiliation{%
  \institution{Lomonosov Moscow State University}
  \city{Moscow}
  \country{Russian Federation}}
\email{egor.chistov@graphics.cs.msu.ru}

\author{Mikhail Voronin}
\orcid{0009-0008-0647-3649}
\affiliation{%
  \institution{Lomonosov Moscow State University}
  \city{Moscow}
  \country{Russian Federation}}
\email{mikhail.voronin@graphics.cs.msu.ru}

\author{Evgeney Bogatyrev}
\orcid{0000-0002-6173-3561}
\affiliation{%
  \institution{MSU Institute for Artificial Intelligence}
  \city{Moscow}
  \country{Russian Federation}}
\email{evgeney.zimin@graphics.cs.msu.ru}

\author{Dmitriy Vatolin}
\orcid{0000-0002-8893-9340}
\affiliation{%
  \institution{MSU Institute for Artificial Intelligence}
  \city{Moscow}
  \country{Russian Federation}}
\email{dmitriy@graphics.cs.msu.ru}

\renewcommand{\shortauthors}{Nikita Alutis et al.}

\begin{abstract}
As video capture moves to handheld and edge devices, motion blur from camera shake has become a pervasive degradation that lowers perceptual quality and harms downstream vision tasks. The strongest deblurring networks recover impressive detail, yet they remain computationally heavy and overwhelmingly complex, so their quality comes at a cost that consumer hardware cannot pay in real time. This gap between restoration quality and on-device speed is exactly what makes real-time deblurring difficult.

We developed and implemented TSRN-RTVD, an efficient video deblurring system that explicitly reconstructs the underlying camera trajectory during exposure and uses the recovered motion to guide restoration. This approach turns the physical cause of blur into a signal that drives sharpening. Our system runs on a single consumer GPU and restores the video at 30 FPS while reaching 30.08 dB PSNR on the GoPro dataset. We demonstrate TSRN-RTVD on consumer devices with interactive side-by-side visualization of the blurry input and the deblurred output, live throughput, and an on-screen view of the recovered camera trajectory. Demo video is available at \url{https://youtu.be/3alMwVrVALU}.
\end{abstract}




\begin{teaserfigure}
  \centering
  \includegraphics[width=\linewidth]{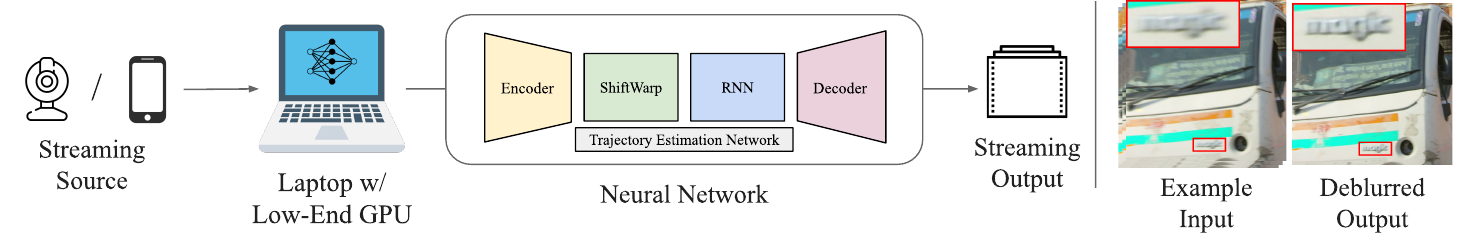}
  \Description{TSRN-RTVD results on a real-blur dataset unseen during training.}
  \caption{System overview of TSRN-RTVD. The method processes a streaming blurry input frame by frame, predicts camera trajectory, performs trajectory-guided recurrent fusion, and outputs deblurred frames in real time. The example is taken from a dataset that was not used for training, demonstrating generalization to real camera motion blur.}
  \label{fig:teaser}
\end{teaserfigure}

\maketitle

\section{Introduction and System Features}

Motion blur is common in handheld and on-board video, where camera shake and fast motion degrade both visual quality and downstream perception. Efficient deblurring techniques are therefore needed to overcome blur and restore sharp videos. Real-time deblurring is especially important for live preview, video streaming, teleoperation, tracking, and detection, where restored frames must be produced as the video arrives. Existing video deblurring methods achieve strong quality on benchmarks such as DVD~\cite{su2017dvd} and GoPro~\cite{nah2017deep}, but many high-quality systems rely on dense optical flow, deformable alignment, recurrent propagation, or long temporal contexts~\cite{edvr, basicvsrpp, dstnet}. These components are effective for offline restoration, but they increase computation and latency in an interactive real-time system. Moreover, due to computational complexity of existing methods, they cannot be run on consumer devices such as phones or laptops, limiting their application.

On the other hand, real-time video deblurring methods do exist~\cite{ifirnn, realtimevdblr, uhdvd}, leveraging intra-frame iterations, multi-patch hierarchy, and spatio-temporal attention. They typically operate in the range of 40--100 FPS, but their restoration quality remains below that of offline methods. This leaves an underexplored region on the speed--quality Pareto frontier, where a different design trade-off is needed. This region is where near-offline restoration quality has to be delivered within a strict per-frame budget, and it motivates the design we describe next.

We present \method{}, a Trajectory-Shift Recurrent Network for Real-Time Video Deblurring system. This system restores $720$p video at $30$ FPS on Apple Silicon M5 using a short three-frame window with one-frame look-ahead. Instead of estimating the dense optical flow between blurry frames or its representations, \method{} uses a compact trajectory signal that approximates the dominant camera-induced motion. Unlike prior real-time designs that spend their compute budget on intra-frame iterations or attention, \method{} spends it on explicit motion, which keeps alignment nearly free and leaves capacity for the fusion and reconstruction stages that actually drive quality.

\begin{figure}[t]
    \centering
    \includegraphics[width=\linewidth]{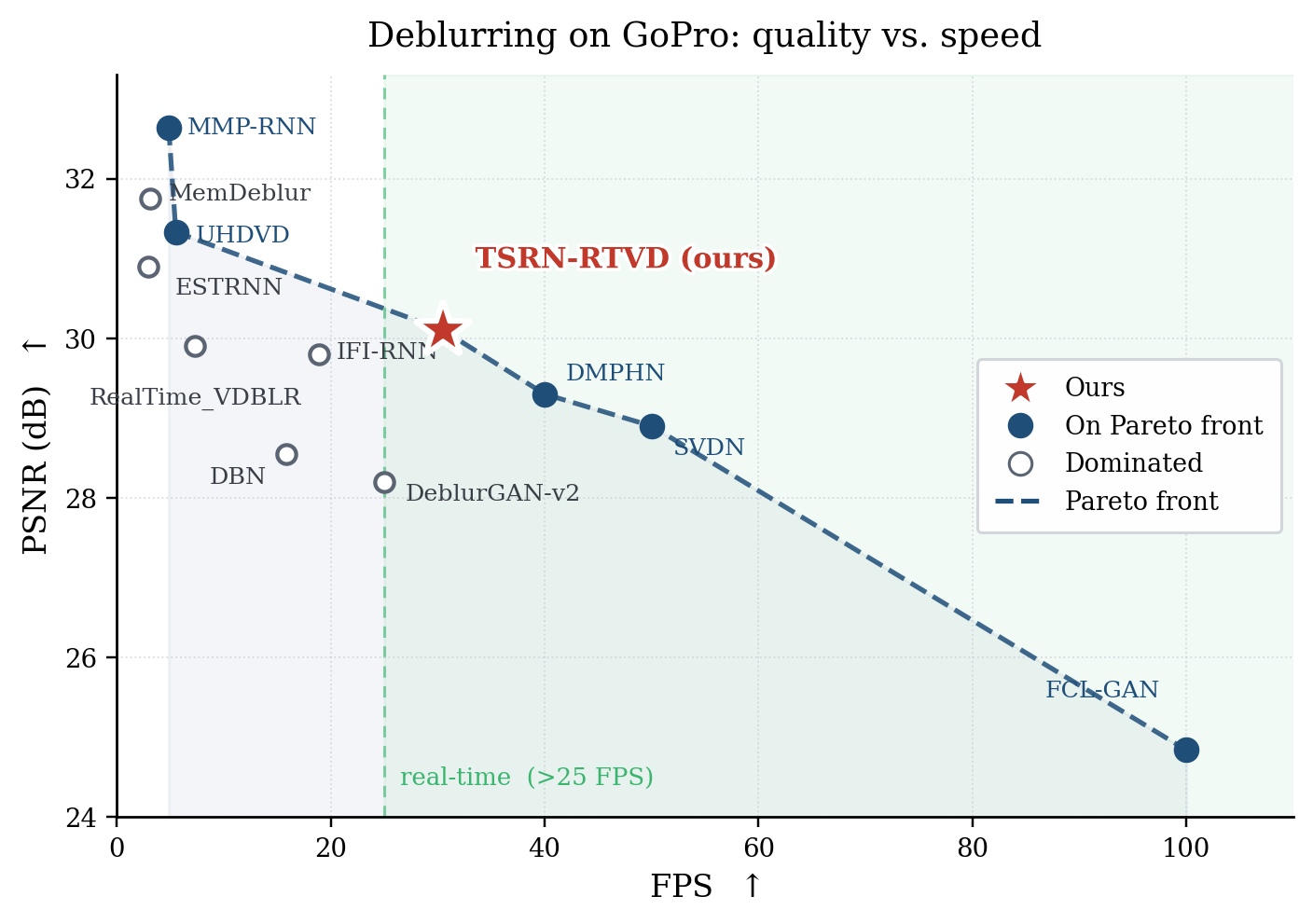}
    \Description{Scatter plot of PSNR versus FPS for GoPro video deblurring methods.}
    \caption{Speed--quality trade-off on GoPro. \method{} operates in the real-time regime and lies on the Pareto frontier between faster low-quality and slower high-quality methods.}
    \label{fig:comparison}
\end{figure}

\section{System Overview}

Figure~\ref{fig:teaser} shows the \method{} pipeline. The system receives a blurry video stream and processes it with a one-frame look-ahead. For each output frame, it uses a three-frame window together with predicted trajectories between neighboring frames. The output is a restored center frame.

\textbf{Trajectory prediction.}
The trajectory represents the projection of the motion of the camera onto the image plane. In practice, it is a compact displacement signal that describes the dominant shift of image content between neighboring frames. During inference, \method{} obtains this signal from a lightweight trajectory-prediction network applied directly to the input video.

\textbf{Trajectory-guided alignment.}
The deblurring network is a compact recurrent residual model operating mostly in low-resolution feature space. A shared encoder extracts features from the current and neighboring blurry frames. The predicted trajectory is then used to shift neighboring-frame features and the recurrent state toward the current frame. This trajectory-guided shift provides a cheap approximation of temporal alignment without dense optical flow, cost volumes, or deformable convolution.

\textbf{Temporal fusion and reconstruction.}
The aligned temporal features, current-frame features, and motion embeddings are fused by convolutional blocks. Rather than imposing a hand-crafted memory rule, the network learns from these inputs how much past and neighboring information to use under different motion patterns. The decoder predicts a residual image which is added to the blurry input frame to produce the restored output.

\textbf{Real-time processing.}
The system is implemented as a streaming pipeline. Encoded features are cached and reused across neighboring windows, so each new frame is encoded once.

\section{Experimental Results}

We evaluate \method{} on the GoPro benchmark~\cite{nah2017deep}, the standard dataset for dynamic-scene motion deblurring. We report restoration quality with PSNR and measure throughput in frames per second. All experiments run at $1280{\times}720$ resolution, which is one of the most common formats for handheld and streaming video, on a single consumer GPU RTX 2080 Ti. Following MIMO-UNet~\cite{cho2021rethinking}, we place a CUDA synchronization barrier around each forward pass, so the measured latency reflects the full GPU workload instead of asynchronous kernel launches.

Figure~\ref{fig:comparison} reports the speed--quality trade-off against streaming and real-time-oriented baselines. We set $25$ FPS as the real-time threshold and compare the methods that produce restored frames as the stream arrives. \method{} reaches $30.08$ dB PSNR while running at $30$ FPS, which places it on the Pareto frontier inside the real-time regime. \method{} fills the previously underexplored region of the frontier and combines real-time speed with restoration quality close to that of slower offline systems.

\section{System Demonstration}
A visitor connects a phone or a webcam to a laptop with a low-end GPU and points it at a scene. The system reads the live stream, runs the full trajectory-shift recurrent pipeline frame by frame, and renders the restored video with no perceptible delay. The interface shows the blurry input and the deblurred output side by side, so visitors compare restoration quality directly on their own footage. We invite visitors to shake the camera or move it quickly, which produces real handheld motion blur, and the system removes it on the fly. The interface also reports live FPS so visitors confirm that processing stays in the real-time regime on consumer hardware.

\section{Conclusion}
We presented \method{}, a real-time trajectory-guided video deblurring system that restores $720$p video at $30$ FPS with a one-frame look-ahead. The system combines lightweight trajectory prediction with compact recurrent restoration, using image-plane camera-motion shifts to align temporal features without dense optical flow. This design keeps the pipeline simple, low-latency, and suitable for interactive video applications.

\begin{acks}
The research was carried out using the MSU-270 supercomputer of Lomonosov Moscow State University.
\end{acks}

\pagebreak

\bibliographystyle{ACM-Reference-Format}
\bibliography{ours}

@INPROCEEDINGS{su2017dvd,
  author={Su, Shuochen and Delbracio, Mauricio and Wang, Jue and Sapiro, Guillermo and Heidrich, Wolfgang and Wang, Oliver},
  booktitle={2017 IEEE Conference on Computer Vision and Pattern Recognition (CVPR)}, 
  title={Deep Video Deblurring for Hand-Held Cameras}, 
  year={2017},
  volume={},
  number={},
  pages={237-246},
  doi={10.1109/CVPR.2017.33}}

@INPROCEEDINGS{nah2017deep,
  author={Nah, Seungjun and Kim, Tae Hyun and Lee, Kyoung Mu},
  booktitle={2017 IEEE Conference on Computer Vision and Pattern Recognition (CVPR)}, 
  title={Deep Multi-scale Convolutional Neural Network for Dynamic Scene Deblurring}, 
  year={2017},
  volume={},
  number={},
  pages={257-265},
  doi={10.1109/CVPR.2017.35}}

@INPROCEEDINGS{edvr,
  author={Wang, Xintao and Chan, Kelvin C.K. and Yu, Ke and Dong, Chao and Loy, Chen Change},
  booktitle={2019 IEEE/CVF Conference on Computer Vision and Pattern Recognition Workshops (CVPRW)}, 
  title={EDVR: Video Restoration With Enhanced Deformable Convolutional Networks}, 
  year={2019},
  volume={},
  number={},
  pages={1954-1963},
  doi={10.1109/CVPRW.2019.00247}}

@INPROCEEDINGS{basicvsrpp,
  author={Chan, Kelvin C.K. and Zhou, Shangchen and Xu, Xiangyu and Loy, Chen Change},
  booktitle={2022 IEEE/CVF Conference on Computer Vision and Pattern Recognition (CVPR)}, 
  title={BasicVSR++: Improving Video Super-Resolution with Enhanced Propagation and Alignment}, 
  year={2022},
  volume={},
  number={},
  pages={5962-5971},
  doi={10.1109/CVPR52688.2022.00588}}

@INPROCEEDINGS{dstnet,
  author={Pan, Jinshan and Xu, Boming and Dong, Jiangxin and Ge, Jianjun and Tang, Jinhui},
  booktitle={2023 IEEE/CVF Conference on Computer Vision and Pattern Recognition (CVPR)}, 
  title={Deep Discriminative Spatial and Temporal Network for Efficient Video Deblurring}, 
  year={2023},
  volume={},
  number={},
  pages={22191-22200},
  doi={10.1109/CVPR52729.2023.02125}}

@INPROCEEDINGS{ifirnn,
  author={Nah, Seungjun and Son, Sanghyun and Lee, Kyoung Mu},
  booktitle={2019 IEEE/CVF Conference on Computer Vision and Pattern Recognition (CVPR)}, 
  title={Recurrent Neural Networks With Intra-Frame Iterations for Video Deblurring}, 
  year={2019},
  volume={},
  number={},
  pages={8094-8103},
  doi={10.1109/CVPR.2019.00829}}

@article{realtimevdblr,
author = {Son, Hyeongseok and Lee, Junyong and Cho, Sunghyun and Lee, Seungyong},
title = {Real-Time Video Deblurring via Lightweight Motion Compensation},
journal = {Computer Graphics Forum},
volume = {41},
number = {7},
pages = {177-188},
doi = {10.1111/cgf.14667},
url = {https://onlinelibrary.wiley.com/doi/abs/10.1111/cgf.14667},
year = {2022}
}

@INPROCEEDINGS{uhdvd,
  author={Deng, Senyou and Ren, Wenqi and Yan, Yanyang and Wang, Tao and Song, Fenglong and Cao, Xiaochun},
  booktitle={2021 IEEE/CVF International Conference on Computer Vision (ICCV)}, 
  title={Multi-Scale Separable Network for Ultra-High-Definition Video Deblurring}, 
  year={2021},
  volume={},
  number={},
  pages={14010-14019},
  doi={10.1109/ICCV48922.2021.01377}}

@INPROCEEDINGS{cho2021rethinking,
  author={Cho, Sung-Jin and Ji, Seo-Won and Hong, Jun-Pyo and Jung, Seung-Won and Ko, Sung-Jea},
  booktitle={2021 IEEE/CVF International Conference on Computer Vision (ICCV)}, 
  title={Rethinking Coarse-to-Fine Approach in Single Image Deblurring}, 
  year={2021},
  volume={},
  number={},
  pages={4621-4630},
  doi={10.1109/ICCV48922.2021.00460}}

\end{document}